# Ovarian Cancer Prediction from Ovarian Cysts Based on TVUS Using Machine Learning Algorithms

Laboni Akter and Nasrin Akhter

Department of Biomedical Engineering, Khulna University of Engineering
& Technology, Khulna, Bangladesh

Emails:laboni.kuet.bme@gmail.com and nasrindr07@gmail.com

**Abstract.** Ovarian Cancer (OC) is type of female reproductive malignancy which can be found among young girls and mostly the women in their fertile or reproductive. There are few number of cysts are dangerous and may it cause cancer. So, it is very important to predict and it can be from different types of screening are used for this detection using Transvaginal Ultrasonography (TVUS) screening. In this research, we employed an actual datasets called PLCO with TVUS screening and three machine learning (ML) techniques, respectively Random Forest KNN, and XGBoost within three target variables. We obtained a best performance from this algorithms as far as accuracy, recall, f1 score and precision with the approximations of 99.50%, 99.50%, 99.49% and 99.50% individually. The AUC score of 99.87%, 98.97% and 99.88% are observed in these Random Forest, KNN and XGB algorithms .This approach helps assist physicians and suspects in identifying ovarian risks early on, reducing ovarian malignancy-related complications and deaths.

## 1   Introduction

Ovarian cancer (OC) has been the world largest seventh leading causes of death and disability in women [1]. In the year 2018, 295,414 women were diagnosed with OC, and 184,799 women died as a result of the condition [2].Ovarian cancer is a disease of the ovaries, the woman reproductive structures that generate eggs and generate estrogen and progesterone. Ovarian cancer treatment is developing, and the best results are usually shown whenever the cancer is detected early.  Epithelial ovarian cancers (EOC) or OC are the most typical types of ovarian cancer [3]. Ovarian cysts are fluid-filled sacs or pockets that form within or on the ovarian membrane. Despite the fact that postmenopausal females have a higher chance of malignancy to premenopausal women, the most of ovarian cysts in postmenopausal women with no unequivocal malignancy markers, such as solid regions, papillary features, or thick uneven members to develop, are benign [4]. The growing utilization ultrasonography in ovarian screenings and new progress in image analysis have boosted the detection of ovarian cysts in untreated post - menopausal females.

Analyzing a vast number of data acquired via several actual patients' databases yields a wealth of data for providing high-quality healthcare at lower prices. The key to lowering the fatality rate from ovarian cancer is early identification. The physicians will use an efficient and trustworthy screening technique to make an initial prognosis. Conventional diagnostic approaches for ovarian cancer include serum cancer antigen 125 (CA-125) screening and transvaginal ultrasonography (TVUS). The significance of categorizing ovarian cancer patients between low - and high categories has prompted numerous healthcare and bioinformatics research groups to investigate the use of machine learning (ML) technologies. As a result, these methods have been used to model the progression and therapy of malignant diseases. Furthermore, the capability of ML algorithms to find essential variables in complicated dataset demonstrates their value. In cancer research, a range of these strategies, such as KNN, Random Forest, and XGBoost, have been widely used to construct prediction algorithms resulting in efficient and precise making decisions.

In this study, the main approach is that many researchers have done a lot of work on ovarian cancer but no author has done the work of predicting ovarian cancer from cysts using ML. So it can be said that this work is the principal to predict ovarian cancer from ovarian cysts. From an unprocessed set of data, this paper applies machine learning algorithms to create an understanding method for early ovarian cancer diagnosis.

## 2    Related Work

Yasodha et al. [5] analyzed big datasets to develop an experience and understanding method of OC for earlier diagnosis. They had used three algorithms for this work that was Multiclass SVM, ANN, and Naïve Bayes. To properly categorize data either normal or abnormal, PGSO is utilized to improve the rough set feature minimization. The performance for SVM, ANN Naïve Bayes of accuracy were 98%, 95%, 93%, specificity 96.7%, 92.9%, 91.4%,  sensitivity 99%, 97%, 96% respectively. Guan et al. [6] approached to predict OC from metabolomics liquid chromatography spectrometry data SVM was used. The SVM algorithm being tested on LC/TOF MS metabolomics data, with the goal of identifying pairings of putative metabolic diagnostic biomarkers. With 90% accuracy, 37 OC patients. Alqudah et al. [7] classified a evaluation of ML and feature selection systems for OC utilizing serum proteome profiling and wavelet features. There are 207 no cancers and 262 ovarian cancers in the given dataset. With 44 features, they employed ANN, SVM, KNN, and ELM ML methods. The accuracy 99%, 99.45% sensitivity 93.21% precision which combining PCA with SVM. Lu et al. [8] performed to predict ovarian cancer using three machine learning algorithms which are DT, LR, and ROMA. They used 235 patients' data where 89 BOT and 146 OC for DT model and 114 patients where 89 BOT and 25 OC for ROMA and LR model with 49 variables. The training data gained the accuracies 79.6%, 87.2%, 84.7% for DT, LR, ROMA algorithms and test data gained 92.1%, 95.6%, 97.4% DT, LR, ROMA algo-rithms respectively. For training data highest for sensitivity DT was 82.2% and specificity 100% for LR whereas for test data the sensitivity 100% for ROMA and specificity 97.8% for LR. Wang et al. [9] accomplished the HE4 seems significant to identifying OC, particularly in the post - menopausal community, according to a meta-analysis consisting on 32 reports that looked at the prognostic significance of HE4, CA125, and ROMA. ROMA and CA125 remain better tests for detecting OC in post - menopausal women. Zhang et al. [10] developed the dual markers that indicated the quantity of epidemiological data in the onset also progress of OC, therefore a linear multi-marker system incorporating CA125, HE4, estradiol, and progesterone was developed. While associated to CA125 or HE4, their multi-marker approach was much better at distinguishing BPM from EOC patients. Robert Chen et al. [11] designed a model of earlier demise in individuals with left-testis malignant tumor with accuracy of 76.1 percent AUC 0.621, sensibility 0.130, positive 0.659 and F1 score 0.216. In this work clinical variables were obtained since a unit of 273 ovarian cancer patients with phase I and II and a ML algorithm for L2 Regression was developed because of number of patients with mortality forecast issue under 20 months, the twenty-fifth percentile of total survival.

## 3  Methodology

The methodology part has been divided into several sections

- Data Collection
- Data Preprocessing
- Imbalanced Data Handling by SMOTE Analysis
- Dataset Splitting
- Feature Scaling
- Machine Learning Algorithm for Classification\
- Model Result and
- Performance Evaluation Methods

The work's procedure flow-diagram is shown in Figure 1.

### 3.1  Data Collection

The data collected in this research work from PLCO dataset of National Cancer Institute (NCI), United States [12]. This dataset has a number of attributes that aren't required for this purpose. As a result, have selected eighteen separate features from the dataset, as well as a parameter with a target. In this work, taken a number of approaches which are numcystl, numcystr, ovary_diaml, ovary_diamr, ovary_voll, ovary_volr, ovcyst_diaml, ovcyst_diamr, ovcyst_morphl, ovcyst_morphr, ovcyst_outlinel, ovcyst_outliner, ovcyst_solidl, ovcyst_solidr, ovcyst_suml, ovcyst_sumr, ovcyst_voll, ovcyst_volr, "ovar_result" as the class.

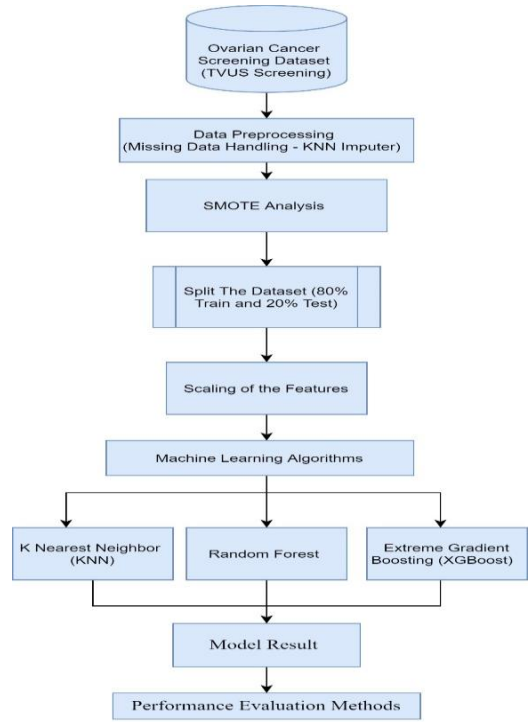

**Fig. 1.** Flowchart of Methodology

### 3.2 Data Preprocessing

*Missing Data Handling by KNN Imputer*
In this dataset there have lots of missing value. First of all, we have selected a feature that has the lowest missing data value. However, a small amount of missing data remains. So, then we used KNN Imputer. Several techniques are available to counter missing values in a dataset. In this research work, the KNN imputer has been applied. In scikit-learn, KNN Imputer is a popular approach for imputing missing data. It is largely accepted as a viable alternative to typical impute methods. Inhere k=5 number of neighbors to eliminate missing data contained in the dataset [13].

### 3.3 Imbalanced Data Handling by SMOTE Analysis

We found data imbalances for several categories when doing the categorization function. Due to the fact that this was an actual medical datasets, unbalanced classifications were unavoidable. When faced with unbalanced datasets, traditional machine learning system analysis methods fail to accurately characterize proposed system. SMOTE (synthetic minority oversampling method) is one of the most often used oversampling approaches for dealing with the imbalanced class problem [14].Fig.2 shows the class wise imbalanced data before the SMOTE and the data set has balanced after SMOTE.

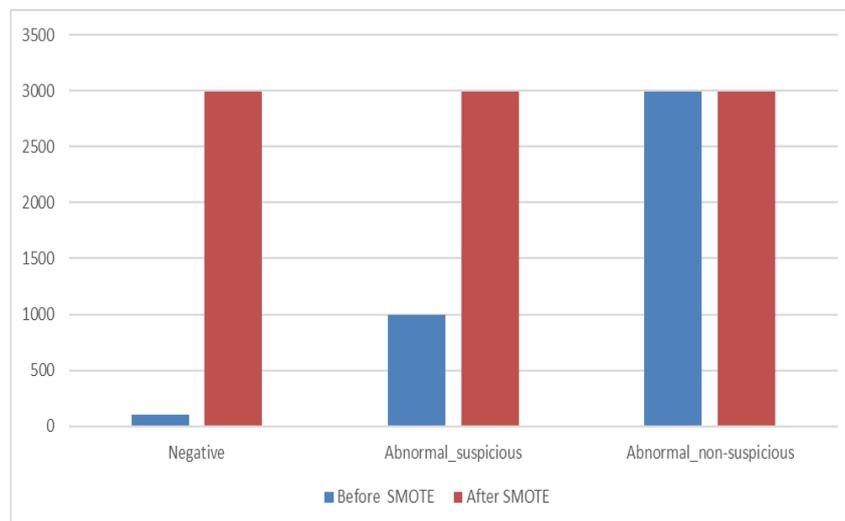

**Fig. 2.** Dataset distributions by class before and after SMOTE

### 3.4 Dataset Splitting

The dataset was split into binary parts: a training set and a test set. The training dataset contained 80% of the whole data, while the test set contained 20% of the total data. The ML establishes a relationship with the independent and dependent parameters in order to foresee or choose an alternative, and then the test data is used to determine whether the ML approach is effective [15].

### 3.5 Feature Scaling

Feature scaling is a technique for bringing most of the features to the same scale. We used min-max feature scaling (normalization) for all of the features in this study. It's a rescaling approach in which estimations are shifted and resized till they're someplace between 0 and 1. Therefore, it's known as normalization. This is a method for normalizing the level of data's self-sufficient features. It is usually done as part of the data pretreatment procedure [16]. The Standard Scalar Python Library was applied in this research.

### 3.6 Implemented Machine Learning Algorithms

**KNN (K Nearest Neighbor)**
The KNN classifier is a common basic ML method which is being utilized to categorize images. It is dependent on the feature vector separation, and we have labeled data to categorize and recover the image's exact class. The Euclidean distance was employed as the similarity function. Model performance is linked to determining the optimal number of neighbors. KNN had a neighbor number of 9 in this study [17].

**Random Forest**
Because of classification, Random Forest (RF) is a supervised ML technique, is used. We concluded that a forest is made up of trees, and that the additional trees there are, the more powerful the forest. Similarly, the RF algorithm proposes DT on data tests and then receives the urge to every one of these before selecting the optimum configuration using voting form projection procedures. It is a way of dressing that reduces overfitting by averaging the results [14].

**XGBoost (Extreme Gradient Boosting)**
XGBoost is a DT based ensemble ML approach that uses gradient boosting. In unorganized dataset forecasting, ANN outperform all established techniques or systems. The XGBoost algorithm was used to do classifying in the dataset. We split the data using percent split methodology, with 80% of the data in the training data and 20% of the data in the test data, and have used classification algorithms. On classification problems, XGBoost works effectively on small data sets. Boosting is a grouping strategy in which newer versions are introduced to resolve current models' combination [16].

### 3.7 Performance Evaluation Methods

The ML outcomes depending on confusion matrix findings were analyzed to determine the efficiency of the deployed ML methods. Accuracy, recall, f1 score and precision are the performance metrics. Four arithmetical keys, like false positive (FP), false negative (FN), true positive (TP), and true negative (TN) were generated to compute the accuracy, precision, recall, and f1 score of this methodology. The accuracy, recall, precision, and f1 score were calculated as follows:

$$Accuracy = (TP+TN) / (TP+TN+FP+FN) \tag{1}$$

$$Recall = TP / (TP+FN) \tag{2}$$

$$Precision = TP / (TP+FP) \tag{3}$$

$$F1score = 2 \times (Precision \times Recall) / (Precision + Recall) \tag{4}$$

## 4 Result and Discussion

Through this portion, we analyze and evaluate the outcomes obtained by the ML methods. Fig. 3 shows the feature correlation matrix. This matrix calculates the correlation across two features in order to reveal their link. The correlation value runs ranging from -1 to +1, with +/-1 denoting negative/positive correlation is a optimal and 0 denoting no connection at all. The diagonal components of this symmetrical matrix are all +1. We clearly see a strong positive correlation between the numcystr and a strong negative connection between the ovary_voll in the matrix. That means that the ovarian cancer class is heavily impacted by the characteristics.

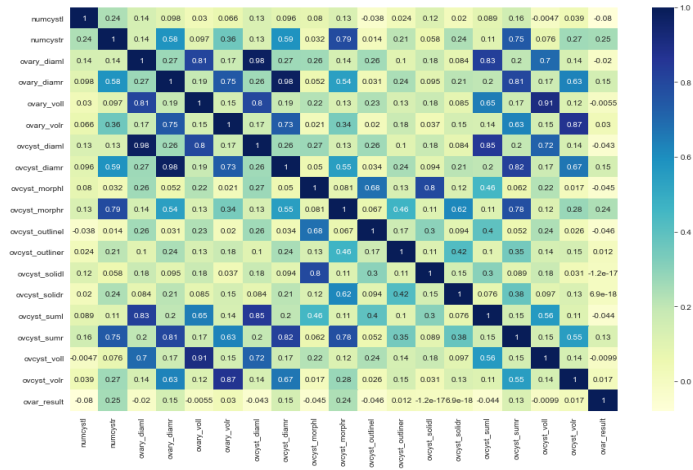

**Fig. 3.** Feature Correlation Matrix

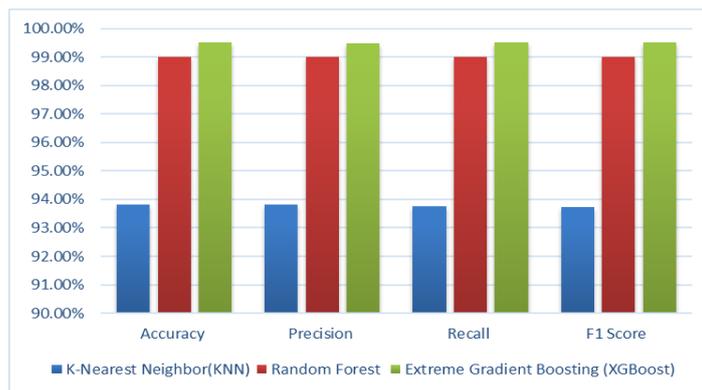

**Fig. 4.** Accuracy, Precision, Recall and F1 score of KNN, Random Forest and XGBoost Algorithms

Fig. 4 shows the results of KNN, Random Forest, XGBoost algorithms for predict the ovarian cancer with three classes "Negative", "Abnormal, suspicious", "Abnormal, non-suspicious". The accuracy of the KNN, Random Forest, and XGBoost were 93.82%, 99%, 99.50% respectively. The precision was gained of KNN, Random Forest, and XGBoost were 93.83%, 98.99%, 99.49% respectively. The recall values which was achieved of KNN, Forest, and XGBoost 93.75%, 99.01%, 99.50% respectively. The F1 Score were obtained of KNN, Random Forest, and XGBoost were 93.73%, 98.99%, 99.50%.

Table 1 shows the comparison between proposed work and the previous study of some papers to predict the ovarian cancer. From this table we can see that some of the author's was not calculated the precision, recall and f1 score and the number of features was also a lot where in this work the number of features is less and the accuracy is high than others.

**Table 1.** Comparison Table with Existing Work

| References | Number of features | Algorithms | Accuracy | Precision | Recall | F1-Score |
|---|---|---|---|---|---|---|
| [6] | N/A | Multiclass SVM ANN NB | 98% | 96.7% | 99% | N/A |
| [7] | N/A | SVM | 90% | N/A | N/A | N/A |
| [8] | 44 | ANN SVM KNN | 99% | 93.21% | 99.45% | N/A |
| [9] | 49 | ROMA DT RF | 84.7% (Train) 97.4% (Test) | 100% | 82.2% | N/A |
| Proposed Result | 18 | KNN RF XGBoost | 99.50% | 99.49% | 99.50% | 99.50% |

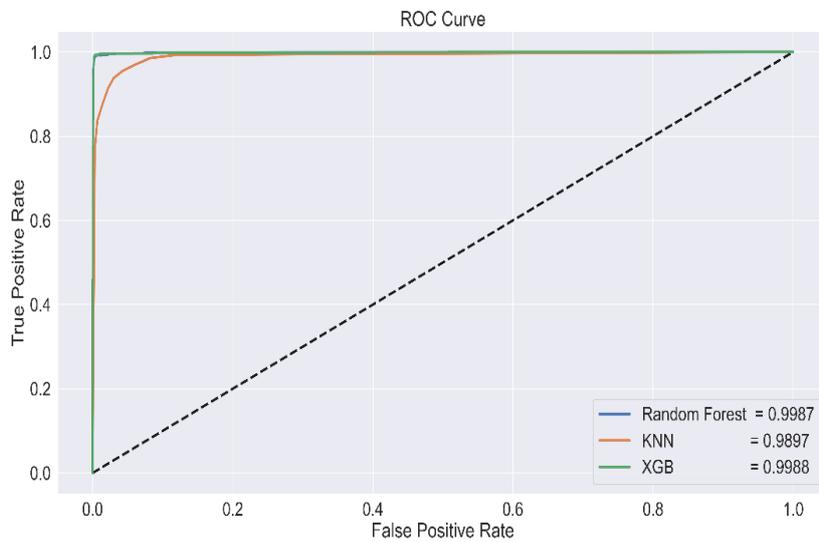

**Fig. 5.** The ROC Curve of Random Forest, KNN, XGBoost Algorithms

We utilize the Area Under the Curve (AUC) and the Receiver Operating Characteristics (ROC) curve to graphically examine the outcomes of the three classes categorization. AUC is a measurement of distinction over classifications obtained by a given classifier while ROC is a probability curve. In this situation, the algorithms performed well in terms of categorization accuracy. The ROC and AUC for the KNN, Random Forest, and XGBoost method in categorizing the three classes are shown in Fig.5. In the three classes' organizations, for Random Forest, KNN, XGBoost the AUC score was determined to be 99.87%, 98.97%, 99.88% correspondingly.

## 5      Conclusion

In this study, we dispensation ML techniques a forecast of ovarian cancer from ovarian cysts of TVUS screening. We obtained high accuracy 99.50%, f1 score 99.50%, recall 99.50% , precision 99.49%  from this KNN, RF, XGBoost algorithms respectively with "Negative", "Abnormal, suspicious", "Abnormal, non-suspicious" classes. In this work, the missing is handle by KNN Imputer and the imbalanced data is handle by SMOTE. The use of as an approach will result in a much more precise assessment of system forecasting accuracy.  Further study will combine the suggested approach with additional techniques like as ultrasonic imaging recognition and merge all utilizing machine learning and deep learning approaches to improve selection process effectiveness. The implementation of this work for early detection of ovarian cancer may beneficial of this deadly disease.